\documentclass[10pt,twocolumn,letterpaper]{article}

\usepackage{wacv}              

\usepackage{amsmath}
\usepackage{amssymb}
\usepackage{amsfonts}
\usepackage{mathtools}
\usepackage{bm}
\usepackage{bbm}
\usepackage{amsthm}

\usepackage{graphicx}
\usepackage{subcaption}
\usepackage{wrapfig}
\usepackage{float}

\usepackage{pifont}

\usepackage{booktabs}
\usepackage{multirow}
\usepackage{makecell}
\usepackage{array}
\usepackage{tabularx}
\usepackage{adjustbox}
\usepackage{colortbl}
\usepackage[table]{xcolor}

\usepackage{enumitem}
\usepackage{xspace}
\usepackage{microtype}
\usepackage{nicefrac}
\usepackage{siunitx}

\usepackage{amsthm}
\theoremstyle{definition}

\usepackage[normalem]{ulem}

\usepackage{amsthm}

\definecolor{wacvblue}{rgb}{0.21,0.49,0.74}
\usepackage[pagebackref,breaklinks,colorlinks,allcolors=wacvblue]{hyperref}

\def\wacvPaperID{1673} 
\def\confName{WACV}
\def\confYear{2027}

\title{PROVE: Probabilistic Visual Grounding via Patch-Level Evidence for 
Vision-Language Models}

\author{Mayank Nautiyal$^{1,2}$ \quad Tianru Zhang$^{1}$ \quad Li Ju$^{1,2}$ \quad Ekta Vats$^{1}$ \quad Prashant Singh$^{1,2}$\\
$^{1}$Department of Information Technology, Uppsala University, Sweden\\
$^{2}$Science for Life Laboratory, Uppsala University, Sweden\\
{\tt\small \{mayank.nautiyal, tianru.zhang, li.ju, ekta.vats, prashant.singh\}@it.uu.se}
}

\begin{document}
\maketitle
\begin{abstract}
Visual grounding with vision-language models such as CLIP typically produces a
single deterministic relevance map for an image-query pair, without indicating
whether the localisation is reliable or whether the referent is present at all. We
introduce \textbf{PROVE} (\textbf{Pro}babilistic \textbf{V}isual Grounding via
Patch-Level \textbf{E}vidence), a post-hoc evidential adapter that maps frozen CLIP
patch features to per-patch Beta distributions without modifying the backbone.
PROVE casts grounding as per-patch membership: each patch accumulates positive and
negative evidence for whether it belongs to the referent, so a query may match
zero, one, or many patches. We train PROVE with an evidential objective on soft
patch-coverage targets, and define a prior-pull objective that keeps absent-referent
examples near an uninformative prior, so that vacuity serves as an absence signal
without a dedicated presence classifier. During inference, a single forward pass
yields, in closed form, a posterior-mean grounding map, a spatial vacuity map, and
an image-level absence score. We evaluate PROVE on referring-expression benchmarks: 
it is competitive in localisation with state-of-the-art CLIP grounding methods, its uncertainty tracks localisation error and supports selective prediction under abstention, and its image-level score detects when the referent is absent altogether. Our evidential patch-membership modelling offers a simple, drop-in mechanism for probabilistic verification in frozen vision-language grounding systems.
\end{abstract}

\begin{figure}[t]
    \centering
    \includegraphics[width=\columnwidth]{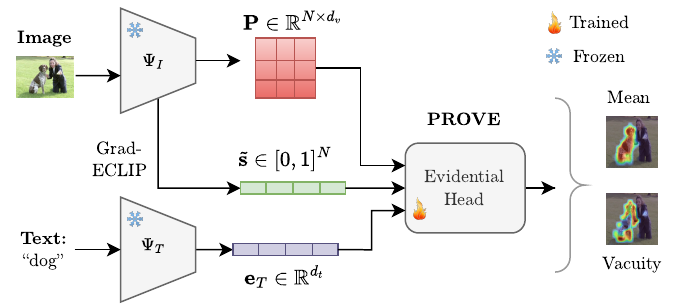}
    \caption{\textbf{Overview of PROVE.} Frozen CLIP encoders ($\Psi_I$, $\Psi_T$) give
        patch and text features and a saliency prior (Grad-ECLIP); a trained evidential head
        maps them to per-patch Betas, giving a mean grounding map and a vacuity map.}
    \label{fig:method_overview}
\end{figure}

\section{Introduction}
\label{sec:introduction}
Vision-language models (VLMs) such as CLIP~\cite{clip} align visual and textual
representations through contrastive training on web-scale image--text pairs,
achieving strong zero-shot performance across classification, retrieval, visual
question answering, and captioning. This alignment is learned only at the global image--text level, yet CLIP's image encoder preserves spatial structure in its intermediate features before they are pooled into a single embedding. The match is
therefore spatially opaque by default: given an image--text pair, the model
exposes a single similarity score but does not convey which image regions support it, limiting interpretation, failure diagnosis, and safe deployment~\cite{Pantazopoulos2025TowardsUV}.

This spatial structure can be leveraged to address the limitations above without retraining.
Rather than training a dedicated end-to-end grounding network from region-level
annotations, one can keep the pretrained backbone frozen and read a relevance map
directly off its patch features, attributing the global image--text match back to
the regions that drive it. Post-hoc methods built on CLIP, such as
Grad-CAM~\cite{selvaraju2017gradcam}, Grad-ECLIP~\cite{zhao2024gradeclip},
attention rollout~\cite{abnar2020quantifying}, and
CLIP-Surgery~\cite{li2023clipsurgery}, take this route, and the resulting maps
assist downstream uses from interpreting an image--text match, to guiding
classifiers and weakly supervised localisation~\cite{gals}, to letting embodied
agents act on objects named in instructions. Each, however, returns only a single
deterministic map, leaving two questions relevant to the user unanswered: how
reliable is the predicted localisation, and is the referent present in the image
at all?

A deterministic map cannot indicate whether its localisation rests on sufficient evidence, and it must place relevance somewhere even when the queried referent is absent, so a salient peak may reflect either a valid localisation or an unsupported response. 
CLIP's contrastive objective brings matched image--text pairs together and pushes mismatched ones apart, but never learns to represent a referent absent from the scene it otherwise matches, so a frozen map has no way to express that nothing was found. Because this gap lies in what the frozen features represent rather than in how they are read, surfacing it motivates adapting the model. 

We close this gap by formulating visual grounding as probabilistic patch
membership. Keeping the per-patch, frozen-backbone setting, we train a lightweight
adapter that replaces each patch's point score with a membership decision: what
fraction of the patch belongs to the referent. We model relevance as membership
rather than allocation, since a referent may occupy no patches, a single patch, or
an extended region, so relevance need not be normalised across spatial locations. This is a Bernoulli question per patch, whose conjugate prior is the Beta distribution. A per-patch
Beta turns the point score into a distribution whose two parameters accumulate
evidence for and against membership above an uninformative $\mathrm{Beta}(1,1)$
prior: where the referent is present, evidence accumulates and the posterior
departs from the prior; where it is absent, no patch departs and the model stays
uncertain everywhere. The posterior mean then gives the grounding score and the
total evidence its vacuity, so localisation and its reliability come from a single
object rather than a heuristic bolted onto a point estimate.

We realise this as \textbf{PROVE} (\textbf{Pro}babilistic \textbf{V}isual
Grounding via Patch-Level \textbf{E}vidence; Fig.~\ref{fig:method_overview}), a post-hoc adapter over frozen CLIP features. From a single forward
pass it produces, in closed form, three readouts of the same per-patch posterior:
a mean grounding map for localisation, a vacuity map flagging where the
prediction is poorly supported, and an image-level absence score. A prior-pull
objective holds no-target examples near the prior, so absence registers as the
lack of any high-evidence patch, requiring no dedicated presence classifier.

\vspace{0.25cm}
\noindent\textbf{Contributions.}
\begin{itemize}
\item We recast CLIP visual grounding as probabilistic patch membership through \textbf{PROVE}, a lightweight post-hoc adapter that predicts a
per-patch Beta posterior over a frozen backbone, yielding localisation,
spatial reliability, and absence detection from a single shared
representation.

\item We derive an evidential training objective from the Beta--Bernoulli
conjugate structure, combining fractional patch-coverage supervision, an
unsupported-evidence regulariser, and a prior-pull term that yields
absent-referent detection without a supervised presence classifier.

\item In both in-distribution and transfer settings, PROVE stays competitive in
localisation with established CLIP grounding methods, while its vacuity ranks
localisation error and supports selective prediction; the proposed prior-pull
objective additionally flags absent referents.
\end{itemize}

\section{Related Work}
\label{sec:related_work}

\noindent\textbf{Visual Grounding and Explanations for VLMs.}
CLIP~\cite{clip} provides a natural foundation for visual grounding because it is already trained on large-scale image--text data and its ViT image encoder exposes spatial patch features in addition to a global image representation. This makes it possible to localise a text query by extracting relevance directly from the frozen visual features, without training an end-to-end grounding model from region-level supervision. Existing attempts to do this often repurpose popular explanation methods. Grad-CAM~\cite{selvaraju2017gradcam}, adapted to CLIP ViTs, uses gradients to weight spatial features, but tends to produce coarse maps because patch tokens contribute to the image--text similarity only indirectly. Attention rollout~\cite{abnar2020quantifying} can be extended to the CLIP ViT by propagating attention across transformer layers, but it is not inherently conditioned on the query and inherits the sparse attention patterns of the encoder.

Methods designed specifically for CLIP generally provide stronger localisation. CLIP-Surgery~\cite{li2023clipsurgery} modifies the inference path to suppress spurious similarity between background patches and the query, while CLIP-ES~\cite{Lin_2023_CVPR} applies softmax-based suppression using an explicit background category set for weakly supervised segmentation. gScoreCAM~\cite{chen2022gscoreCAM} selects the top-k feature channels with the highest gradient magnitudes to reduce noise and complexity, and Grad-ECLIP~\cite{zhao2024gradeclip} combines gradient-derived channel weights with loosened spatial attention to produce sharper query-conditioned maps. Despite their different mechanisms, these methods return a single deterministic relevance map for each image--query pair. They therefore provide no explicit measure of localisation reliability and cannot distinguish a well-supported match from an unsupported response when the queried referent is absent.

A related line improves grounding by finetuning a full vision-language backbone end-to-end. AMC~\cite{yang2023improving} finetunes ALBEF (Align Before Fuse)~\cite{li2021albef}  with a margin loss that aligns its GradCAM explanations with human region annotations, and SelfEQ~\cite{he2024selfEQ} finetunes ALBEF for self-consistent explanation maps under LLM-generated paraphrase supervision. These methods update the entire backbone on a different VLM and obtain strong localisation, but at a heavier training cost and without any measure of predictive reliability. PROVE targets a strictly lighter, backbone-controlled setting by keeping CLIP entirely frozen and training only a small post-hoc adapter. It further differs in two ways. First, rather than emitting a point map, it predicts a per-patch Beta distribution, so localisation and its reliability come from a single object. Second, it takes the Grad-ECLIP saliency prior as input and refines it into an evidential map with per-patch uncertainty. The same evidential structure lets it reserve a low-evidence state for absent referents, whereas the suppression heuristics in CLIP-ES~\cite{Lin_2023_CVPR} treat absence only as a background problem to be filtered by hand.

\vspace{0.25cm}
\noindent\textbf{Uncertainty in vision--language models.}
Uncertainty in VLMs has been studied, but mostly at the global image-text level: keep CLIP frozen and add a lightweight post-hoc component over its embeddings to model a distribution, differing mainly in mechanism and in whether they target aleatoric or epistemic uncertainty~\cite{upadhyay2023probvlm, baumann2024post, ju2025exploiting, ju2026epistemicuncertaintyquantificationpretrained, venkataramanan2025probabilistic}. This global uncertainty is suited to retrieval, where the question is whether an image and a query are compatible at all. Grounding entails a more specific question: not whether the query matches, but where, and how reliable that localisation is. A single pooled embedding or similarity score cannot capture which patches drive the match or separate reliable localisations from unsupported ones. PROVE keeps CLIP frozen and trains a lightweight head, but places uncertainty directly on patch membership, yielding spatially resolved relevance and vacuity.

\vspace{0.25cm}
\noindent\textbf{Beta-based evidential uncertainty.} Beta distributions are a natural way to model uncertainty over binary probabilities and bounded quantities. Rather than predicting a single value, an evidential model outputs the two Beta parameters, whose mean gives the prediction and whose concentration reflects the amount of accumulated evidence, so the estimate and its uncertainty are obtained from a single forward pass~\cite{sensoy2018evidential, malinin2018prior, bezirganyan2025emsec}. This formulation has been used in multi-label classification, where each label is assigned a Beta distribution for out-of-distribution detection~\cite{aguilar2025mlood}; in dense prediction, where per-pixel Beta distributions provide spatial uncertainty maps for segmentation and anomaly detection~\cite{kim2025distributional}; and in bounded regression, where the spread of a per-example Beta captures predictive uncertainty~\cite{elallaf2026betarisk}. PROVE adapts this formulation to language-conditioned grounding by assigning a Beta distribution to the membership of each image patch in the queried region. The resulting evidence indicates where the referent is supported, while a lack of accumulated evidence across all patches signals that the referent is absent.

\section{Method}
\label{sec:method}

\subsection{Problem Setup and Overview}
\label{subsec:overview}
Given an image $I$ and a referring expression $T$, a frozen CLIP
ViT~\cite{clip} partitions the image into $N=L^2$ patches on an $L\times L$ grid
and encodes each patch as a feature $\mathbf{p}_i\in\mathbb{R}^{d_v}$, collected
into $\mathbf{P}=\{\mathbf{p}_i\}_{i=1}^{N}$. The query is encoded as a
unit-normalised text embedding $\mathbf{e}_T\in\mathbb{R}^{d_t}$, with $d_v$ and
$d_t$ fixed by the backbone. From the same frozen image-text pair we additionally
extract a deterministic patch-level localisation cue
$\tilde{\mathbf{s}}=(\tilde{s}_1,\ldots,\tilde{s}_N)\in[0,1]^N$ using
Grad-ECLIP~\cite{zhao2024gradeclip}. All CLIP and Grad-ECLIP parameters remain
frozen.

We formulate visual grounding as patchwise probabilistic membership estimation. For
each patch $i$ we introduce a relevance variable $r_i\in[0,1]$ representing the
fraction of the patch that belongs to the region described by $T$: values near $1$
indicate membership, near $0$ background, and intermediate values partial coverage
at the patch resolution. Rather than predicting a single relevance score, PROVE
parameterises a predictive distribution
$q_\phi(r_i\mid\mathbf{P},\mathbf{e}_T,\tilde{\mathbf{s}})$ for each patch,
separating the expected relevance used for localisation from the uncertainty behind
it and letting the model flag unreliable localisations and absent-referent queries.
Because a referent may occupy no patches, a single patch, or an extended region, we
do not normalise relevance across spatial locations, avoiding the artificial
competition induced by a simplex constraint.

\begin{figure}[t]
    \centering
    \includegraphics[width=\columnwidth]{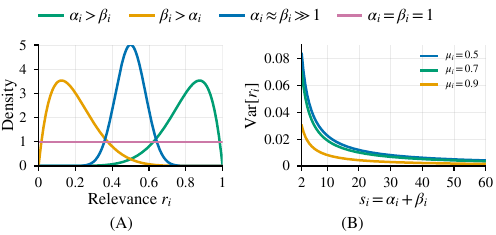}
    \caption{\textbf{Beta evidence and predictive uncertainty.} (a) The balance of
    $\alpha_i$ and $\beta_i$ shifts relevance toward foreground or background;
    $\alpha_i\!\approx\!\beta_i\!\gg\!1$ concentrates around intermediate relevance, and
    $\alpha_i=\beta_i=1$ is the uniform low-evidence prior. (b) For fixed mean $\mu_i$,
    higher concentration $s_i=\alpha_i+\beta_i$ reduces the predictive variance.}
    \label{fig:beta_relevance}
    \label{fig:beta_relevance}
\end{figure}

\subsection{Patchwise Beta Relevance}
\label{subsec:beta}
Whether a patch belongs to the referent is a Bernoulli event; we let
$r_i\in[0,1]$ be its membership probability. The Beta distribution is the conjugate
prior of the Bernoulli likelihood, making it the natural model for $r_i$: its
parameters accumulate positive and negative evidence on top of the uninformative
reference $\mathrm{Beta}(1,1)$, 
\begin{equation}
    r_i \sim \mathrm{Beta}(\alpha_i, \beta_i),
    \qquad \alpha_i, \beta_i \geq 1,
    \qquad i = 1, \ldots, N.
    \label{eq:beta}
\end{equation}
Treating patches as conditionally independent given the parameters, the relevance
field $\mathbf{r}=(r_1,\ldots,r_N)$ factorises into $N$ independent Betas on
$[0,1]^N$,
\begin{equation}
    p(\mathbf{r}\mid\boldsymbol{\alpha},\boldsymbol{\beta})
    = \prod_{i=1}^{N}
      \frac{r_i^{\,\alpha_i-1}(1-r_i)^{\,\beta_i-1}}{B(\alpha_i,\beta_i)},
    \label{eq:beta_joint}
\end{equation}
where $B(\alpha_i,\beta_i)=\Gamma(\alpha_i)\Gamma(\beta_i)/\Gamma(\alpha_i+\beta_i)$
is the Beta function.

\subsection{Patch-Level Predictive Quantities}
\label{subsec:summaries}
Each predicted Beta distribution provides three complementary patch-level
quantities: the mean, the vacuity, and the variance. The mean,
\begin{equation}
    \mu_i = \frac{\alpha_i}{s_i},
    \qquad
    s_i = \alpha_i + \beta_i,
    \label{eq:mean}
\end{equation}
serves as the grounding score. Evidence in $\alpha_i$ shifts the mean toward membership, whereas evidence in $\beta_i$ shifts it toward background. At the uniform prior, $\alpha_i=\beta_i=1$, the mean is $\mu_i=0.5$ but no evidence has been accumulated (Fig.~\ref{fig:beta_relevance}(a)). We quantify this lack of evidence using the vacuity ~\cite{josang2018uncertainty},
\begin{equation}
    u_i = \frac{2}{s_i} \in (0,1],
    \label{eq:vacuity}
\end{equation}
which equals one at the prior and decreases as evidence accumulates in either
direction. The predictive variance,
\begin{equation}
    \mathrm{Var}[r_i]
    =
    \frac{\mu_i(1-\mu_i)}{s_i+1}
    \label{eq:variance}
\end{equation}
measures the spread of the relevance variable around its mean and decreases
monotonically with $s_i$ for fixed $\mu_i$
(Fig.~\ref{fig:beta_relevance}(b)). Together, these quantities distinguish predictions with similar mean relevance
but different evidential support. For example, $\mu_i\approx0.5$ may arise from
an uninformed prediction with $s_i\approx2$ and high vacuity, or from a
well-supported intermediate relevance estimate with large, balanced evidence.
We use these summaries for grounding, uncertainty estimation, and
absent-referent detection in Sec.~\ref{subsec:inference}.

\subsection{Evidential Parameterisation}
\label{subsec:head}
A learnable head $f_\phi$ maps the frozen image-text inputs to patch-wise Beta parameters. Since $\alpha_i,\beta_i \geq 1$, we represent them as a unit reference plus non-negative evidence, $\alpha_i = 1 + e_i^{+}$ and $\beta_i = 1 + e_i^{-}$,
so the head only needs to predict how much evidence to add for and against membership. For each patch, it outputs two raw values
$\hat{e}_i^{+},\hat{e}_i^{-}$, mapped to bounded non-negative evidence by,
\begin{equation}
    e_i^{\pm}= \gamma \tanh\!\left(
    \frac{\operatorname{softplus}(\hat{e}_i^{\pm})}{\gamma}\right).
    \label{eq:bounded_evidence}
\end{equation}
Zero evidence recovers the reference $\mathrm{Beta}(1,1)$, while larger $e_i^{+}$ or $e_i^{-}$ adds support for relevance or background. The capacity $\gamma$ bounds the evidence, $0 \leq e_i^{\pm} < \gamma$, so $1 \leq \alpha_i,\beta_i < 1+\gamma$, preventing unbounded concentration; it fixes the otherwise underdetermined evidence scale, since the supervision
constrains the mean relevance but not the concentration. We treat $\gamma$ as a capacity parameter and study its effect empirically in Sec.~S5 of the supplementary.

\subsection{Training Objective}
\label{subsec:objective}
We supervise the relevance variables on the patch grid. For ground-truth region mask $\mathcal{M}$, let $\Omega_i$ be the image area of patch $i$; the patch target is the fractional coverage,
\begin{equation}
    y_i = \frac{|\mathcal{M} \cap \Omega_i|}{|\Omega_i|} \in [0,1].
\end{equation}
Interior patches have $y_i\approx 1$, background patches $y_i\approx 0$, and boundary patches intermediate values. Fractional targets match the support of $r_i\in[0,1]$, preserve sub-patch spatial coverage, and avoid the quantisation introduced by binary patch labels.


\paragraph{Localisation.}
Since $r_i$ is a random variable, we minimise the expected binary cross-entropy
under the predicted Beta,
\begin{equation}
\mathcal{L}_{\mathrm{loc},i}
=
\mathbb{E}_{r_i\sim\mathrm{Beta}(\alpha_i,\beta_i)}
\bigl[-y_i\log r_i - (1-y_i)\log(1-r_i)\bigr].
\label{eq:loc_loss_expectation}
\end{equation}
Using $\mathbb{E}[\log r_i]=\psi(\alpha_i)-\psi(s_i)$ and
$\mathbb{E}[\log(1-r_i)]=\psi(\beta_i)-\psi(s_i)$, with $s_i=\alpha_i+\beta_i$ and
$\psi$ the digamma function, this reduces to,
\begin{equation}
\mathcal{L}_{\mathrm{loc},i}
=
y_i\bigl(\psi(s_i)-\psi(\alpha_i)\bigr)
+
(1-y_i)\bigl(\psi(s_i)-\psi(\beta_i)\bigr).
\label{eq:loc_loss_closed}
\end{equation}
This aligns the mean $\mu_i=\alpha_i/s_i$ with the coverage target (encouraging
$\alpha_i>\beta_i$ for foreground patches and the reverse for background), and for a
fixed mean it decreases as the concentration $s_i$ grows, so it also drives evidence
upward.


\paragraph{Unsupported evidence regulariser.}
Because the localisation loss rewards higher concentration, evidence can accumulate
even in the channel the target does not support: for a partially covered patch both
$\alpha_i$ and $\beta_i$ grow. The capacity $\gamma$ (Sec.~\ref{subsec:head}) bounds
the total concentration, and to ensure evidence accrues only where the target
supports it we penalise deviation from $\mathrm{Beta}(1,1)$ on the unsupported
channel,
\begin{equation}
\begin{aligned}
\mathcal{L}_{\mathrm{reg},i}
=\;& (1-y_i)\,\mathrm{KL}\!\left[\mathrm{Beta}(\alpha_i,1)\,\|\,\mathrm{Beta}(1,1)\right] \\
   & + y_i\,\mathrm{KL}\!\left[\mathrm{Beta}(1,\beta_i)\,\|\,\mathrm{Beta}(1,1)\right].
\end{aligned}
\label{eq:wrong_channel_loss}
\end{equation}
where $\mathrm{KL}[\,\cdot\,\|\,\cdot\,]$ is the Kullback-Leibler divergence. The weighting by $y_i$ and $1-y_i$ is the natural soft extension of the standard
evidential regulariser~\cite{sensoy2018evidential}: covered portions suppress
background evidence and background portions suppress membership evidence, while the
supported channel remains free to accumulate.

\paragraph{Absence.}
When the query has no referent, no mask is projected. Every patch is background, but
training these patches with confident background evidence ($\beta_i\gg\alpha_i$)
would make an absent image indistinguishable from the background patches surrounding
a present referent, where the same confident-background state occurs. To keep global
absence identifiable from patch evidence, we reserve the low-evidence prior state for
no-target pairs and pull every patch toward the reference,
\begin{equation}
\mathcal{L}_{\mathrm{nt},i}
=
\mathrm{KL}\!\left[\mathrm{Beta}(\alpha_i,\beta_i)\,\|\,\mathrm{Beta}(1,1)\right].
\label{eq:notarget_loss}
\end{equation}
An absent referent then yields $\alpha_i\approx\beta_i\approx 1$ everywhere (high
vacuity), whereas a present referent retains at least one high-evidence patch, the
signal the absence score exploits (Sec.~\ref{subsec:inference}).

\paragraph{Full objective.}
Let $z\in\{0,1\}$ indicate a present referent and $w_i$ a class-balanced patch
weight that upweights the rarer foreground patches. We
average the per-patch terms,
\begin{equation}
\mathcal{L}
=
\frac{z}{N}\sum_{i=1}^{N} \bigl(w_i\,\mathcal{L}_{\mathrm{loc},i}+\lambda(t)\,\mathcal{L}_{\mathrm{reg},i}\bigr)
+
\frac{(1-z)\,\lambda_{\mathrm{nt}}}{N}\sum_{i=1}^{N}\mathcal{L}_{\mathrm{nt},i},
\label{eq:final_loss}
\end{equation}
where $\lambda(t)$ is annealed linearly from $0$ to $\lambda_{\mathrm{reg}}$ over the
first training steps, so localisation is learned before confidence is
shaped~\cite{sensoy2018evidential}; $\lambda_{\mathrm{reg}}\ge 0$ is the final weight
of the unsupported-evidence regulariser, and $\lambda_{\mathrm{nt}}\ge 0$ weights the
absence term.

\subsection{Inference, Abstention, and Absence Detection}
\label{subsec:inference}
A single forward pass predicts $(\alpha_i,\beta_i)$ for every patch, from which
the mean relevance and vacuity are obtained in closed form without sampling.
These quantities define the grounding map, provide an image-level measure of
prediction reliability, and support absent-referent detection.

\vspace{0.25cm}
\noindent\textbf{Grounding.}
We use the posterior-mean map
$\hat{\mathbf{m}}=(\mu_1,\ldots,\mu_N)$ as the grounding prediction, with
$\mu_i=\alpha_i/s_i$ for patch $i$. Since $\mu_i$ is trained against fractional
patch coverage, a patch is classified as foreground when $\mu_i>0.5$, indicating
that it is predicted to be more covered than uncovered by the referent.

\vspace{0.25cm}
\noindent\textbf{Image-level uncertainty.}
We summarise the reliability of a grounding prediction by aggregating the
evidence supporting its predicted foreground. Averaging over all patches would
mix evidence for the localisation with evidence for background and therefore
obscure the confidence associated with the predicted region itself. Let
$\mathcal{F}=\{i:\mu_i>0.5\}$ denote the predicted foreground, and define,
\begin{equation}
    U_{\mathrm{fg}}
    =
    \frac{1}{|\mathcal{F}|}
    \sum_{i\in\mathcal{F}} u_i,
    \qquad
    u_i = \frac{2}{s_i}.
    \label{eq:image_vacuity}
\end{equation}
Once the foreground set is selected, $U_{\mathrm{fg}}$ depends only on the
accumulated evidence within that region. Low values indicate a well-supported
localisation, whereas high values indicate weak evidential support; we use
$U_{\mathrm{fg}}$ to rank predictions for selective abstention
(Sec.~\ref{subsec:eval}).

\vspace{0.25cm}
\noindent\textbf{Absence detection.}
The prior-pull objective in Eq.~\eqref{eq:notarget_loss} encourages every patch
of an absent-referent pair to remain near $\mathrm{Beta}(1,1)$, yielding
$s_i\approx2$ throughout the image. In contrast, a present referent should cause
at least one patch to accumulate evidence and depart from the prior. We therefore
define absence by the lack of any high-evidence patch:
\begin{equation}
    A(I,T) = -\max_i s_i.
    \label{eq:absence_inference}
\end{equation}
Larger values of $A(I,T)$ indicate less evidence that the referent is present.
An image--text pair is classified as absent when $A(I,T)\geq\tau$, where
$\tau$ is selected on the validation split and applied unchanged to the test
sets. This score is obtained directly from the predicted patch evidence and
requires no separate presence classifier. We ablate other choices for this
score, such as the membership evidence $\alpha_i$ and a top-$k$ pool over
patches, in Sec.~S4 of the supplementary.

\begin{table*}[!t]
\centering
\caption{\textbf{Grounding localisation.} Best in \textbf{bold}, second best
\uline{underlined}. PROVE: mean$_{\pm\text{std}}$ over 5 seeds; baselines
deterministic.}
\label{tab:main_results}
\renewcommand{\arraystretch}{1.05}
\resizebox{\textwidth}{!}{%
\begin{tabular}{lcccccccccccc}
\toprule
& \multicolumn{3}{c}{\textbf{gRefCOCO (people)}}
& \multicolumn{3}{c}{\textbf{gRefCOCO (objects)}}
& \multicolumn{3}{c}{\textbf{Flickr30k Entities}}
& \multicolumn{3}{c}{\textbf{PhraseCut}} \\
\cmidrule(lr){2-4} \cmidrule(lr){5-7} \cmidrule(lr){8-10} \cmidrule(lr){11-13}
\textbf{Method}
& PG$\uparrow$ & EBPG$\uparrow$ & IoU@$k\uparrow$
& PG$\uparrow$ & EBPG$\uparrow$ & IoU@$k\uparrow$
& PG$\uparrow$ & EBPG$\uparrow$ & IoU@$k\uparrow$
& PG$\uparrow$ & EBPG$\uparrow$ & IoU@$k\uparrow$ \\
\midrule
Grad-CAM~\cite{selvaraju2017gradcam}
& 0.202 & 0.225 & 0.179
& 0.181 & 0.229 & 0.209
& 0.400 & 0.358 & 0.320
& 0.295 & 0.215 & 0.234 \\
Grad-ECLIP~\cite{zhao2024gradeclip}
& \uline{0.552} & \uline{0.372} & 0.283
& \uline{0.522} & \uline{0.358} & 0.302
& \uline{0.820} & \textbf{0.567} & 0.440
& \uline{0.586} & \textbf{0.350} & 0.350 \\
CLIP-Surgery~\cite{li2023clipsurgery}
& 0.495 & 0.263 & \uline{0.299}
& 0.433 & 0.265 & \uline{0.310}
& \textbf{0.822} & 0.400 & \uline{0.457}
& \textbf{0.630} & 0.245 & \textbf{0.403} \\
Rollout~\cite{abnar2020quantifying}
& 0.124 & 0.200 & 0.146
& 0.081 & 0.166 & 0.098
& 0.238 & 0.303 & 0.239
& 0.116 & 0.151 & 0.101 \\
\midrule
\textbf{PROVE (ours)}
& $\mathbf{0.855}_{\pm.001}$ & $\mathbf{0.459}_{\pm.004}$ & $\mathbf{0.616}_{\pm.002}$
& $\mathbf{0.759}_{\pm.005}$ & $\mathbf{0.412}_{\pm.004}$ & $\mathbf{0.538}_{\pm.003}$
& $0.764_{\pm.005}$ & $\underline{0.502}_{\pm.004}$ & $\mathbf{0.527}_{\pm.003}$
& $0.512_{\pm.014}$ & $\underline{0.274}_{\pm.003}$ & $\underline{0.354}_{\pm.011}$ \\
\bottomrule
\end{tabular}}
\end{table*}

\begin{figure*}[!t]
    \centering
    \includegraphics[width=0.8\linewidth]{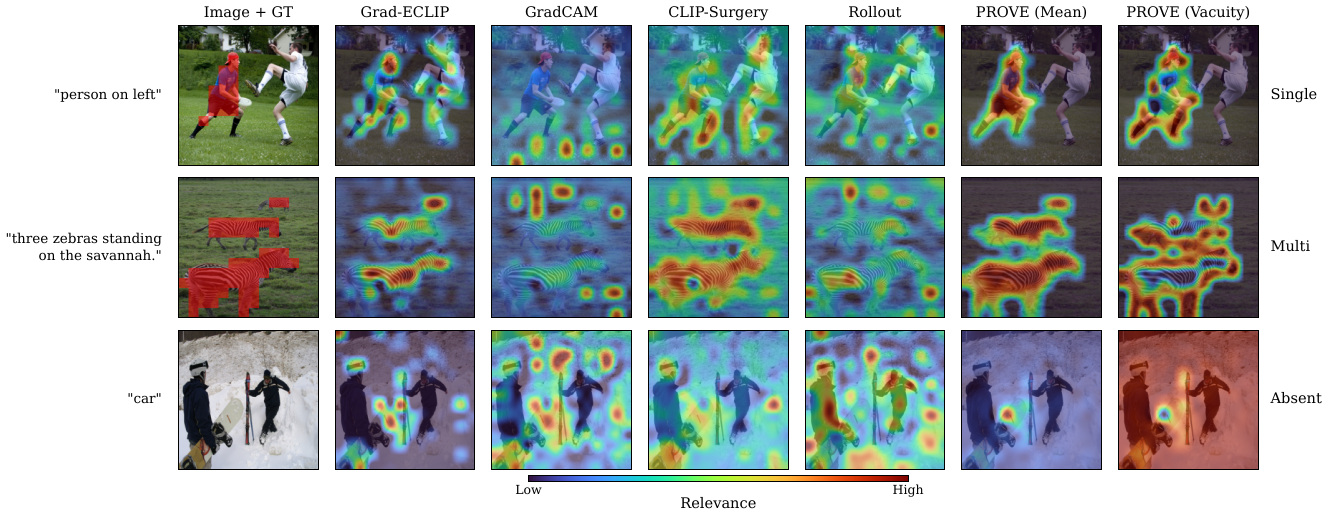}
    \caption{\textbf{Qualitative comparison} on single-, multi-, and no-target queries.
PROVE's mean map localises the referent(s), while vacuity flags the absent case; the
baselines peak even when the referent is absent. Maps are min--max normalised.}
    \label{fig:qualitative}
\end{figure*}

\section{Experiments}
\label{sec:experiments}
\subsection{Experimental Setup}
\label{subsec:setup}
\noindent\textbf{Backbone and features.}
We use a frozen CLIP ViT-B/16~\cite{clip}. All images are resized to
$224\times224$ pixels before encoding. ViT-B/16 uses non-overlapping patches of
size $16\times16$ pixels, producing $224/16=14$ patches along each spatial
dimension and hence a $14\times14$ grid of $N=196$ patch features. Each patch
feature has dimension $d_v=768$, while the text embedding has dimension
$d_t=512$. The patch features, text embedding, and Grad-ECLIP cue introduced in
Sec.~\ref{subsec:overview} are precomputed and cached for each image-text pair, so
training updates only the evidential head. To test whether the evidential head is tied to a specific backbone, we additionally
report results with a different and much larger frozen backbone (ViT-L/14) in
Sec.~S3 of the supplementary.

\vspace{0.25cm}
\noindent\textbf{Data and target construction.}
We train PROVE on gRefCOCO~\cite{grefcoco}, which contains single-target,
multi-target, and no-target expressions over COCO~\cite{lin2014coco} images. Each expression forms
one example. For target-present examples, we merge all annotated referent masks
and compute the fractional coverage of each CLIP patch to obtain soft targets
$y_i$ (Sec.~\ref{subsec:objective}); hard masks
$\mathbb{I}[y_i>0.5]$ are used only for evaluation. The training split contains
$\sim\!209\mathrm{k}$ examples ($57.6\%$ single-target, $33.2\%$ multi-target,
and $9.1\%$ no-target), while the validation split contains
$\sim\!14\mathrm{k}$ examples and is used for model selection and
absence-threshold calibration. We evaluate on gRefCOCO testA
(People, $\sim\!19\mathrm{k}$) and testB
(Objects, $\sim\!16\mathrm{k}$), and assess cross-dataset generalisation on
Flickr30k Entities~\cite{plummer2015flickr30k} and
PhraseCut~\cite{wu2020phrasecut} ($\sim\!10\mathrm{k}$ each). Dataset details are
given in Sec.~S1 of the supplementary.

\vspace{0.25cm}
\noindent\textbf{Architecture.}
We realise the evidential head $f_\phi$ as follows. Each frozen CLIP patch feature
is layer-normalised and projected to $256$ dimensions. We condition the projected
features on the text embedding using FiLM~\cite{perez2018film}, then add embeddings
of the corresponding Grad-ECLIP scores and patch positions. A global token,
constructed from the pooled image features and text embedding, is prepended to the
$N$ patch tokens. The resulting sequence is processed by a two-layer Transformer
encoder with $4$ attention heads, an FFN dimension of $1024$, pre-normalisation,
GELU activations, and dropout $0.1$. Finally, a two-layer MLP maps each patch token
to the raw positive and negative evidence values $(\hat{e}_i^{+},\hat{e}_i^{-})$,
which are converted to $(\alpha_i,\beta_i)$ as described in Sec.~\ref{subsec:head}
using evidence capacity $\gamma=20$. Although the predictive distribution factorises
across patches, the parameters of each patch depend on the full spatial sequence
through self-attention.

\vspace{0.25cm}
\noindent\textbf{Training.}
We train PROVE for $10{,}000$ steps with batch size $512$ using AdamW (lr
$5{\times}10^{-4}$, weight decay $0.01$), $1500$ warm-up steps followed by cosine
decay, and gradient clipping at $5.0$; an EMA with decay $0.999$ is kept for
evaluation. The unsupported-evidence regulariser ($\lambda_{\mathrm{reg}}=2.0$) is
linearly annealed over the first $2500$ steps, and the absence term uses
$\lambda_{\mathrm{nt}}=1.0$. We train five models with different random seeds and
report mean and standard deviation across them.

\subsection{Evaluation Protocol}
\label{subsec:eval}

We structure our evaluation around three research questions.
\textbf{Q1:} Does the evidential mean map match or exceed deterministic grounding baselines?
\textbf{Q2:} Does predictive uncertainty correlate with grounding correctness, and does abstaining on uncertain predictions improve usable accuracy?
\textbf{Q3:} Can the model detect absent referents without a supervised presence classifier?

\paragraph{Q1: Localisation.}
For a test pair $(I,T)$, let $R$ be its ground-truth positive patch set and $M$ the
number of evaluated pairs; all metrics are averaged over the $M$ pairs. The
\textbf{Pointing Game (PG)}~\cite{zhang2018top} scores a hit when the
highest-relevance patch lies inside $R$. \textbf{EBPG}~\cite{wang2020score} is the
fraction of the $\ell_1$-normalised relevance mass that falls in $R$,
$\mathrm{EBPG}=\sum_{i\in R}\tilde{\mu}_i$, where $\tilde{\boldsymbol{\mu}}$ is the
normalised Beta mean map. \textbf{IoU@$k$}~\cite{everingham2010pascal} forms
$\hat{R}$ from the top-$|R|$ patches of $\boldsymbol{\mu}$ and measures overlap,
$\mathrm{IoU@}k=|\hat{R}\cap R|/|\hat{R}\cup R|$. Pairs with an absent referent are
excluded; evaluated separately in Q3.

Table~\ref{tab:main_results} reports the results. On the in-distribution gRefCOCO
splits, PROVE leads on all three metrics, e.g.\ IoU@$k$ of $0.616$ on people versus
$0.299$ for the best baseline and PG of $0.855$ versus $0.552$. This margin should be
read in light of the setting: PROVE is trained on gRefCOCO whereas the baselines are
training-free, so much of the in-distribution gain reflects supervision rather than
the evidential head, which we isolate with $\mathrm{PROVE}^{\mathrm{det}}$ in
Sec.~\ref{sec:ablation}. Under transfer, PROVE generalises without retraining: it
attains the best IoU@$k$ on Flickr30k and remains competitive elsewhere, while the
saliency baselines lead on PG and EBPG on the transfer sets. We report these transfer
differences descriptively, as several of them are small.

\vspace{0.25cm}
\noindent\textbf{Qualitative comparison.}
Figure~\ref{fig:qualitative} compares PROVE with the grounding baselines on single-, multi-, and absent-referent queries. For \emph{``person on left''}, PROVE localises the specified person. For \emph{``three zebras standing on
the savannah''}, the mean map covers all three referents without distributing a
fixed amount of relevance among them, illustrating the benefit of modelling
patch membership without spatial normalisation. In both cases, vacuity is lowest
on the confidently rejected background and higher on and around the referents,
especially at their boundaries. For the absent query \emph{``car''}, the baseline
methods still produce salient peaks, whereas PROVE's mean map shows only a small
residual peak and its vacuity is high almost everywhere. Thus, the vacuity map
distinguishes an unsupported residual response from a well-supported localisation
and provides a signal for rejecting the query as absent. Additional qualitative
examples for each dataset are provided in Sec.~S6 of the supplementary.

\begin{figure}[!t]
    \centering
    \begin{subfigure}{0.48\columnwidth}
        \centering
        \includegraphics[width=\textwidth]{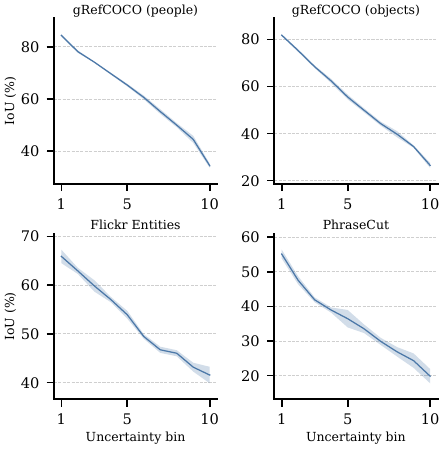}
        \caption{Uncertainty ranking.}
        \label{fig:uncertainty_bins}
    \end{subfigure}
    \hfill
    \begin{subfigure}{0.48\columnwidth}
        \centering
        \includegraphics[width=\textwidth]{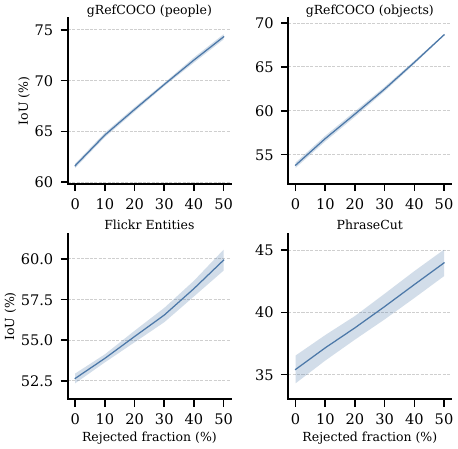}
        \caption{Selective prediction.}
        \label{fig:selective_prediction}
    \end{subfigure}
    \caption{\textbf{Uncertainty as a ranking signal.} (a) IoU@$k$ per vacuity bin
    (bin~1 most certain). (b) IoU@$k$ on the retained set as the most uncertain
    images are rejected. Both span four datasets; mean over 5 seeds, shaded
    $\pm1$ std.}
    \label{fig:q3}
\end{figure}

\paragraph{Q2: Uncertainty as a grounding reliability signal.}
A useful uncertainty score should be higher for image-text pairs that the model localises poorly. We first test whether the image-level vacuity $U_{\mathrm{fg}}$ from Sec.~\ref{subsec:inference} ranks predictions by localisation quality. We sort
the test pairs by $U_{\mathrm{fg}}$ into ten equal-sized bins, from most certain (bin~1) to most uncertain (bin~10), and report the mean IoU@$k$ per bin (Fig.~\ref{fig:uncertainty_bins}). IoU@$k$ decreases \emph{monotonically} with vacuity across all four datasets, with the lowest-vacuity bins localising best and the highest worst and no reversals. This is precisely the behaviour expected of a reliable uncertainty signal; localisation quality falls steadily as predicted uncertainty rises, both in-distribution and under transfer.

Next, we test whether this ranking supports selective grounding. Test pairs are
rejected in decreasing order of $U_{\mathrm{fg}}$, and the mean IoU of the retained
predictions is measured at rejection rates $r\in\{0\%,10\%,\ldots,50\%\}$
(Fig.~\ref{fig:selective_prediction}). Retained-set IoU@$k$ rises
\emph{monotonically} as the most uncertain predictions are abstained on, again following the
ideal trend for an abstention signal. For example, IoU@$k$ on gRefCOCO testA
(people) increases from $\sim\!62\%$ to $\sim\!75\%$ at a $50\%$ rejection rate.



\paragraph{Q3: Absent-referent detection.}
We evaluate whether the absence score $A(I,T)=-\max_i s_i$ from
Sec.~\ref{subsec:inference} can distinguish queries whose referent is absent from
those with a valid localisation; gRefCOCO provides ground-truth labels for both
cases. AUROC measures threshold-free separability, i.e.\ the probability that a
randomly chosen absent pair is scored higher than a present pair ($0.5$ is chance,
$1$ perfect ordering). For a deployable detector we also fix a threshold $\tau$ and
report two operating-point rates,
\begin{equation}
\begin{aligned}
    T\text{-acc} &= \Pr[A(I,T)<\tau \mid \mathrm{present}], \\
    N\text{-acc} &= \Pr[A(I,T)\geq\tau \mid \mathrm{absent}],
\end{aligned}
\label{eq:tnacc}
\end{equation}
where $T\text{-acc}$ is the fraction of present pairs correctly retained (not
abstained on) and $N\text{-acc}$ the fraction of absent pairs correctly rejected. We
select $\tau$ on validation so that at least $90\%$ of present pairs are retained
($T\text{-acc}\geq0.90$), ensuring abstention rarely discards a valid localisation,
and apply it unchanged to both test splits; $N\text{-acc}$ then reports how many
absences are caught at this safe operating point. We compare PROVE with the
Grad-ECLIP score $-\max_i\tilde{s}_i$ and a deterministic control
$-\max_i\mu_i^{\mathrm{det}}$ that uses the same architecture and absence supervision
but replaces the Beta output with a sigmoid, testing whether the evidential
parameterisation adds information beyond the shared model and supervision.

\begin{table}[t]
\centering
\caption{\textbf{Absent-referent detection} on gRefCOCO testA (people) and testB
(objects). Best per split in \textbf{bold}; PROVE is mean$_{\pm\text{std}}$ over 5
seeds, baselines deterministic.}
\label{tab:absence}
\scriptsize
\setlength{\tabcolsep}{3pt}
\renewcommand{\arraystretch}{0.9}
\begin{tabular}{ll cccc}
\toprule
Split & Method & AUROC & N-acc & T-acc & Bal \\
\midrule
\multirow{3}{*}{testA}
 & Grad-ECLIP     & 0.682 & 0.570 & 0.711 & 0.640 \\
 & Det.\ control  & 0.823 & 0.643 & 0.864 & 0.754 \\
 & \textbf{PROVE} & $\mathbf{0.849}_{\pm.010}$ & $\mathbf{0.657}_{\pm.029}$
                  & $\mathbf{0.928}_{\pm.017}$ & $\mathbf{0.792}_{\pm.006}$ \\
\midrule
\multirow{3}{*}{testB}
 & Grad-ECLIP     & 0.746 & 0.518 & 0.853 & 0.686 \\
 & Det.\ control  & 0.789 & \textbf{0.671} & 0.749 & 0.710 \\
 & \textbf{PROVE} & $\mathbf{0.813}_{\pm.006}$ & $0.669_{\pm.026}$
                  & $\mathbf{0.869}_{\pm.021}$ & $\mathbf{0.769}_{\pm.004}$ \\
\bottomrule
\end{tabular}
\end{table}


As shown in Table~\ref{tab:absence}, PROVE achieves the highest AUROC on both splits
($0.849$ on testA, $0.813$ on testB) versus $0.682/0.746$ for Grad-ECLIP and
$0.823/0.789$ for the control. The gain over the control indicates that total
evidence carries a stronger absence signal than deterministic confidence alone. The
validation-selected threshold also transfers: PROVE keeps $T\text{-acc}\geq0.86$
while rejecting about two-thirds of absent queries on both splits.

\section{Ablation Study}
\label{sec:ablation}

\noindent\textbf{A1: Deterministic localisation control.}
The grounding baselines in Table~\ref{tab:main_results} are training-free, whereas
PROVE is trained on localisation supervision, so that comparison does not reveal how
much of PROVE's localisation gain follows simply from supervised adaptation. We
isolate this with $\mathrm{PROVE}^{\mathrm{det}}$, which shares PROVE's post-hoc
architecture up to the output layer: PROVE predicts two evidence values per patch
that parameterise a Beta distribution, whereas $\mathrm{PROVE}^{\mathrm{det}}$
predicts a single logit with a sigmoid, trained with binary cross-entropy on the soft
coverage targets and without the unsupported-evidence regulariser or the prior-pull
objective.

Table~\ref{tab:det_vs_prove} shows that supervised adaptation, not the evidential
head, accounts for the localisation gain over the training-free baselines:
$\mathrm{PROVE}^{\mathrm{det}}$ matches PROVE on PG, is slightly higher on IoU@$k$, and
is clearly higher on EBPG across all four datasets. This trade-off is expected as
$\mathrm{PROVE}^{\mathrm{det}}$ optimises localisation alone, whereas PROVE must
simultaneously fit the coverage targets, organise positive and negative evidence, and
reserve the low-evidence state for absent referents. A model that also represents
uncertainty naturally gives up a small amount of localisation relative to one trained
purely to localise. However, PROVE stays within a small margin of the
deterministic control on PG and IoU@$k$ while providing the vacuity used for
selective grounding (Q2) and a stronger absence signal than the deterministic
control (Q3, Table~\ref{tab:absence}), which may be valuable in several applications.


\vspace{0.25cm}
\noindent\textbf{A2: Loss-term ablation.}
The full objective (Eq.~\eqref{eq:final_loss}) combines the localisation loss
$\mathcal{L}_{\mathrm{loc}}$ with two auxiliary terms: the unsupported-evidence
regulariser $\mathcal{L}_{\mathrm{reg}}$ and the prior-pull loss
$\mathcal{L}_{\mathrm{nt}}$. We measure what each auxiliary term contributes by
removing it on its own, i.e.\ setting its coefficient to zero while keeping
$\mathcal{L}_{\mathrm{loc}}$, the architecture, and all other hyperparameters fixed
(Table~\ref{tab:a1_ablation}); localisation is always trained, and we never remove
more than one term at a time. Removing $\mathcal{L}_{\mathrm{nt}}$ leaves
localisation unchanged or slightly improved but reduces absence detection to near
chance: AUROC falls from $0.83$ to $0.56$ on testA and from $0.80$ to $0.57$ on
testB, and balanced accuracy drops to $0.50$ on both splits. The prior-pull term is
therefore what produces the absence signal, at a negligible localisation cost.
Removing $\mathcal{L}_{\mathrm{reg}}$ barely affects the threshold-free absence
ranking but degrades localisation and the transfer of the validation-selected
threshold: IoU@$k$ drops by about $0.02$ to $0.03$, and $T$-acc falls from $0.95$ to
$0.85$ on testA and from $0.90$ to $0.83$ on testB. Thus $\mathcal{L}_{\mathrm{reg}}$
controls the unsupported evidence, stabilising both localisation and the thresholded
absence decision. In summary, $\mathcal{L}_{\mathrm{nt}}$ creates the absence signal,
while $\mathcal{L}_{\mathrm{reg}}$ regularises its evidence scale and operating point.

\section{Limitations}
\label{sec:limit}
PROVE is a supervised, post-hoc adapter on a frozen backbone; therefore a part of its localisation gain stems from supervised adaptation rather than the evidential parameterisation alone. This is expected, and an acceptable tradeoff: the
supervision is what lets PROVE learn a probabilistic model, and in return it provides the vacuity signal and a stronger absence signal than a purely deterministic adapter. As is common for supervised grounding methods, its capability is bounded under distribution shift; nonetheless, the degradation on the
out-of-distribution Flickr30k and PhraseCut transfer sets is modest, and PROVE remains competitive, indicating that the learned evidential signal transfers reasonably well. Because it conditions on frozen CLIP features and a Grad-ECLIP cue,
PROVE inherits their biases and any localisation errors in the prior, and would benefit directly from stronger backbones or saliency priors. Extending PROVE beyond CLIP to other vision-language backbones such as SigLIP~\cite{zhai2023siglip} is a promising direction, requiring only a backbone-appropriate saliency prior in
place of Grad-ECLIP.

\begin{table}[!t]
\centering
\caption{\textbf{Deterministic vs.\ probabilistic localisation.}
PROVE: mean over 5 seeds. Higher value per metric in \textbf{bold} (PG comparable across both).}
\label{tab:det_vs_prove}
\setlength{\tabcolsep}{3pt}
\resizebox{\columnwidth}{!}{%
\begin{tabular}{l ccc ccc ccc ccc}
\toprule
& \multicolumn{3}{c}{People} & \multicolumn{3}{c}{Objects}
& \multicolumn{3}{c}{Flickr30k} & \multicolumn{3}{c}{PhraseCut} \\
\cmidrule(lr){2-4}\cmidrule(lr){5-7}\cmidrule(lr){8-10}\cmidrule(lr){11-13}
Method & PG & EBPG & IoU@$k$ & PG & EBPG & IoU@$k$
& PG & EBPG & IoU@$k$ & PG & EBPG & IoU@$k$ \\
\midrule
$\mathrm{PROVE}^{\mathrm{det}}$
& 0.86 & \textbf{0.63} & \textbf{0.63}
& 0.76 & \textbf{0.55} & \textbf{0.55}
& 0.77 & \textbf{0.68} & \textbf{0.55}
& 0.52 & \textbf{0.41} & \textbf{0.36} \\
\textbf{PROVE}
& 0.86 & 0.46 & 0.62
& 0.76 & 0.41 & 0.54
& 0.76 & 0.50 & 0.53
& 0.51 & 0.27 & 0.35 \\
\bottomrule
\end{tabular}}
\end{table}

\section{Conclusion}
\label{sec:conclusion}
We presented PROVE, a post-hoc evidential adapter that enables probabilistic grounding from a frozen CLIP backbone, predicting a per-patch Beta distribution over referent membership instead of a single deterministic map. Casting grounding as per-patch
membership lets a query match zero/one/many patches, and grounds the model in the conjugate Beta-Bernoulli structure, where each patch accumulates positive and negative evidence on an uninformative prior. A prior-pull objective keeps
absent-referent examples near this prior, making vacuity an absence signal without a dedicated presence classifier. From a single forward pass, PROVE yields in closed form a grounding map, a spatial uncertainty map, and an image-level absence
score. Across referring-expression benchmarks it is competitive in localisation with state-of-the-art CLIP grounding methods, its uncertainty tracks localisation error and supports selective prediction, and it detects absent referents, outperforming a deterministic control trained with the same supervision and thereby isolating the value of the evidential parameterisation. Since the backbone stays frozen, PROVE is a drop-in addition to any CLIP-based grounding pipeline.

\begin{table}[!t]
\centering
\caption{\textbf{Loss ablation} on gRefCOCO testA (people) and testB (objects), single-seed runs.
\emph{Full} is the default PROVE configuration; each lower row removes one auxiliary
term.}
\label{tab:a1_ablation}
\scriptsize
\setlength{\tabcolsep}{3pt}
\renewcommand{\arraystretch}{0.9}
\begin{tabular}{llccccc}
\toprule
& & \multicolumn{2}{c}{Loc.} & \multicolumn{3}{c}{Absence} \\
\cmidrule(lr){3-4}\cmidrule(lr){5-7}
Split & Variant & PG & IoU@$k$ & AUROC & $T$-acc & Bal. \\
\midrule
\multirow{3}{*}{testA}
& Full                          & 0.85 & 0.61 & 0.83 & 0.95 & 0.78 \\
& $-\mathcal{L}_{\mathrm{nt}}$  & 0.86 & 0.63 & 0.56 & 1.00 & 0.50 \\
& $-\mathcal{L}_{\mathrm{reg}}$ & 0.84 & 0.59 & 0.83 & 0.85 & 0.78 \\
\midrule
\multirow{3}{*}{testB}
& Full                          & 0.76 & 0.54 & 0.80 & 0.90 & 0.76 \\
& $-\mathcal{L}_{\mathrm{nt}}$  & 0.76 & 0.54 & 0.57 & 1.00 & 0.50 \\
& $-\mathcal{L}_{\mathrm{reg}}$ & 0.74 & 0.51 & 0.79 & 0.83 & 0.76 \\
\bottomrule
\end{tabular}
\end{table}

\section*{Acknowledgments}
The experiments and data processing were carried out on the Berzelius and Alvis computing resources. Berzelius is provided by the Knut and Alice Wallenberg Foundation at the National Supercomputer Centre and was used under project Berzelius-2026-100. Alvis is provided by Chalmers e-Commons at Chalmers and was accessed through the National Academic Infrastructure for Supercomputing in Sweden (NAISS) under project NAISS~2025/5-98. NAISS is partially funded by the Swedish Research Council through grant agreement no.~2022-06725. Prashant Singh acknowledges funding from the Swedish Research Council under grant agreement no.~2023-05593. Prashant Singh also acknowledges support from the Knut and Alice Wallenberg Foundation through the Wallenberg AI, Autonomous Systems and Software Program (WASP) and the Data-Driven Life Science (DDLS) NEST project ``TIMED''. Ekta Vats acknowledges support from the Kjell \& M\"{a}rta Beijer Foundation.

\section*{Code Availability}
Code for data preparation, training and all experiments is available at
\url{https://github.com/mayank05942/PROVE}.

{
    \small
    \bibliographystyle{ieeenat_fullname}
    \bibliography{main}
}

\clearpage
\onecolumn

\renewcommand{\thefigure}{S\arabic{figure}}
\renewcommand{\thetable}{S\arabic{table}}
\renewcommand{\thesection}{S\arabic{section}}
\renewcommand{\theequation}{S\arabic{equation}}
\renewcommand{\theHsection}{S\arabic{section}}
\renewcommand{\theHfigure}{S\arabic{figure}}
\renewcommand{\theHtable}{S\arabic{table}}
\renewcommand{\theHequation}{S\arabic{equation}}
\setcounter{figure}{0}
\setcounter{table}{0}
\setcounter{section}{0}
\setcounter{equation}{0}

\setcounter{figure}{0}
\setcounter{table}{0}
\setcounter{section}{0}
\setcounter{equation}{0}
\begin{center}
{\Large\bfseries Supplementary Material}
\end{center}

\section{Datasets}
\label{sec:data}

We construct training and evaluation splits from three publicly available referring expression datasets.

\paragraph{gRefCOCO~\cite{grefcoco}.}
We use the \texttt{FudanCVL/gRefCOCO} HuggingFace release\footnote{\url{https://huggingface.co/datasets/FudanCVL/gRefCOCO}},
which extends referring-expression grounding on COCO images with
single-target, multi-target, and no-target expressions. We take the
referring expressions (\texttt{grefs(unc).json}) together with the COCO
instance annotations (\texttt{instances.json}) from the same repository,
and treat each expression as one example. The \texttt{train} split is our
sole training source, while the \texttt{val} split is used for model
selection and absence-threshold calibration; the \texttt{testA}
(people-centric) and \texttt{testB} (object-centric) splits serve as
in-domain evaluation benchmarks.

\paragraph{Flickr30k Entities ~\cite{plummer2015flickr30k}.}
We use the test images from \texttt{lmms-lab/flickr30k} HuggingFace
release\footnote{\url{https://huggingface.co/datasets/lmms-lab/flickr30k}} paired with
phrase-region annotations from the Flickr30k Entities repository. We retain
only those phrase and region pairs that have valid chain mappings and
referring expressions of at least two words, yielding 9{,}905 evaluation
examples used exclusively for zero-shot transfer evaluation.

\paragraph{PhraseCut ~\cite{wu2020phrasecut}.}
We use the test split of VGPhraseCut (\texttt{refer\_test.json}), retaining
tasks that have valid polygon annotations and successfully loaded images,
yielding 10{,}004 evaluation examples used exclusively for zero-shot
transfer evaluation.

\section{Backbones and Feature Extraction}

All experiments use a \emph{frozen} OpenAI CLIP ViT image--text encoder.
We report our main results with ViT-B/16 and replicate the main findings
with the larger ViT-L/14 (Sec.~\ref{supp:vitl14}) to show the evidential
head is not specific to one backbone. The two share an identical pipeline:
images are resized to $224\times224$, the encoder produces an $L\times L$
grid of $N=L^2$ patch tokens, a Grad-ECLIP saliency prior is computed on the
\emph{same} frozen backbone, and the patch features, text embedding, and
saliency prior are precomputed and cached; only the evidential head is
trained. They differ only in capacity and patch grid: ViT-B/16
($\sim$150M parameters) yields a $14\times14$ grid of $196$ tokens with
patch dimension $d_v\!=\!768$ and text dimension $d_t\!=\!512$, whereas
ViT-L/14 ($\sim$428M) yields a $16\times16$ grid of $256$ tokens with
$d_v\!=\!1024$ and $d_t\!=\!768$. The head adapts to these dimensions
through its input projection and positional embeddings; all other components
are unchanged.

\begin{table}[H]
\centering
\caption{Localisation across backbones on gRefCOCO. PROVE: mean$\pm$std over
five seeds (four for ViT-L/14).}
\label{tab:supp_backbone_loc}
\setlength{\tabcolsep}{4pt}
\renewcommand{\arraystretch}{1.2}
\footnotesize
\begin{tabular}{l l ccc}
\toprule
\textbf{Split} & \textbf{Backbone} & \textbf{PG}$\uparrow$ & \textbf{EBPG}$\uparrow$ & \textbf{IoU@$k$}$\uparrow$ \\
\midrule
\multirow{2}{*}{testA}
  & ViT-B/16 & $0.855{\scriptstyle\,\pm.001}$ & $0.459{\scriptstyle\,\pm.004}$ & $0.616{\scriptstyle\,\pm.002}$ \\
  & ViT-L/14 & $\mathbf{0.871}{\scriptstyle\,\pm.002}$ & $\mathbf{0.479}{\scriptstyle\,\pm.004}$ & $\mathbf{0.632}{\scriptstyle\,\pm.002}$ \\
\addlinespace
\multirow{2}{*}{testB}
  & ViT-B/16 & $0.759{\scriptstyle\,\pm.005}$ & $0.412{\scriptstyle\,\pm.004}$ & $0.538{\scriptstyle\,\pm.003}$ \\
  & ViT-L/14 & $\mathbf{0.773}{\scriptstyle\,\pm.003}$ & $\mathbf{0.433}{\scriptstyle\,\pm.003}$ & $\mathbf{0.560}{\scriptstyle\,\pm.001}$ \\
\bottomrule
\end{tabular}
\end{table}

\section{Results with a Larger Backbone}
\label{supp:vitl14}
\begin{figure}[H]
\centering
\subfloat[]{%
    \includegraphics[width=0.35\linewidth]{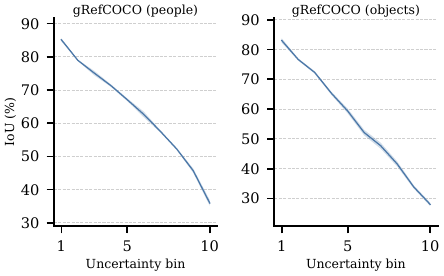}%
    \label{fig:vitl14_uncert_a}
}
\hspace{0.7cm}
\subfloat[]{%
    \includegraphics[width=0.35\linewidth]{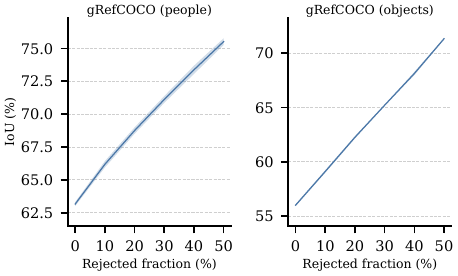}%
    \label{fig:vitl14_uncert_b}
}
\caption{ViT-L/14 uncertainty behaviour on gRefCOCO.
(a) IoU@$k$ across vacuity bins, where bin~1 is the most certain.
(b) IoU@$k$ on the retained set as a function of the rejected fraction.
Mean over five seeds, with shading indicating $\pm1$ standard deviation.}
\label{fig:supp_vitl14_uncert}
\end{figure}

Tables~\ref{tab:supp_backbone_loc} and~\ref{tab:supp_backbone_abs} report
localisation and absent-referent detection on gRefCOCO testA and testB for
both backbones. The main-paper findings carry over and, on localisation,
improve: ViT-L/14 raises PG, EBPG, and IoU@$k$ on both splits
(Table~\ref{tab:supp_backbone_loc}), consistent with PROVE benefiting
directly from a stronger frozen encoder. Absence detection is preserved,
with marginally higher AUROC and balanced accuracy than ViT-B/16
(Table~\ref{tab:supp_backbone_abs}). The uncertainty signal behaves as in the experiments in the main text: IoU@$k$ decreases monotonically with vacuity, and rejecting the
most uncertain predictions raises retained IoU@$k$ (
Fig.~\ref{fig:supp_vitl14_uncert}). This confirms that the evidential
parameterisation, and not a particular backbone drives the result.

\begin{table}[H]
\centering
\caption{Absent-referent detection across backbones on gRefCOCO,
$A=-\max_i s_i$. PROVE: mean$\pm$std over five seeds (four for ViT-L/14).}
\label{tab:supp_backbone_abs}
\setlength{\tabcolsep}{4pt}
\renewcommand{\arraystretch}{1.2}
\footnotesize
\begin{tabular}{l l cccc}
\toprule
\textbf{Split} & \textbf{Backbone} & \textbf{AUROC} & \textbf{$N$-acc} & \textbf{$T$-acc} & \textbf{Bal.} \\
\midrule
\multirow{2}{*}{testA}
  & ViT-B/16 & $0.849{\scriptstyle\,\pm.010}$ & $0.657{\scriptstyle\,\pm.029}$ & $0.928{\scriptstyle\,\pm.017}$ & $0.792{\scriptstyle\,\pm.006}$ \\
  & ViT-L/14 & $\mathbf{0.851}{\scriptstyle\,\pm.008}$ & $\mathbf{0.660}{\scriptstyle\,\pm.020}$ & $\mathbf{0.936}{\scriptstyle\,\pm.012}$ & $\mathbf{0.798}{\scriptstyle\,\pm.004}$ \\
\addlinespace
\multirow{2}{*}{testB}
  & ViT-B/16 & $0.813{\scriptstyle\,\pm.006}$ & $\mathbf{0.669}{\scriptstyle\,\pm.026}$ & $0.869{\scriptstyle\,\pm.021}$ & $0.769{\scriptstyle\,\pm.004}$ \\
  & ViT-L/14 & $\mathbf{0.822}{\scriptstyle\,\pm.008}$ & $0.667{\scriptstyle\,\pm.016}$ & $\mathbf{0.888}{\scriptstyle\,\pm.013}$ & $\mathbf{0.778}{\scriptstyle\,\pm.003}$ \\
\bottomrule
\end{tabular}
\end{table}

\section{Absence Score Design}
\label{supp:absence}

We score absence by the total evidence,
\begin{equation}
    A(I,T) = -\max_i s_i,
    \qquad s_i = \alpha_i + \beta_i.
    \label{eq:abs_strength}
\end{equation}
We justify this choice against two alternatives, evaluated post-hoc on the same trained models. Training shapes the two evidence channels asymmetrically. A foreground patch
of a present pair concentrates its evidence in $\alpha_i$ while the regulariser
holds $\beta_i\!\approx\!1$, so $s_i\!\approx\!\alpha_i$; a background patch does
the reverse, with large $\beta_i$ and $\alpha_i\!\approx\!1$; and every patch of
an absent pair is drawn back to the prior, $\alpha_i\!\approx\!\beta_i\!\approx\!1$
and $s_i\!\approx\!2$. Consequently a high strength can arise from either channel
in a present image, a high membership $\alpha_i$ only from the foreground, and an
absent image is low in both.

This explains why \eqref{eq:abs_strength} is the appropriate statistic. The strength
$-\max_i s_i$ reports an image as present as soon as any patch is confident in
either channel, and because the prior-pull keeps absent images low, even a small
or weakly-localised referent is still detected: its surroundings remain
confidently background, $\beta_i$ is large, and $s_i$ stays high. A
membership-only score,
\begin{equation}
    A_\alpha(I,T) = -\max_i \alpha_i,
    \label{eq:abs_alpha}
\end{equation}

matches the semantic notion of presence and is immune to background leakage, but
it discards the background channel; present images whose foreground evidence is
weak are then pushed toward the absent cluster and fewer are retained. A top-$k$
pooled strength,
\begin{equation}
    A_{k}(I,T) = -\frac{1}{k}\sum_{i\in\mathrm{top}\text{-}k} s_i,
    \label{eq:abs_topk}
\end{equation}

replaces the maximum with the mean of the $k$ largest strengths and tests whether
the signal depends on a single patch.

Table~\ref{tab:supp_absence} reports AUROC and the thresholded operating point
for the three scores. The strength score~\eqref{eq:abs_strength} attains the best
AUROC and the highest $T$-acc: the background channel rescues weakly-localised
present cases, while the prior-pull keeps absent images low so the added
sensitivity costs no false positives. The membership-only
score~\eqref{eq:abs_alpha} trails on $T$-acc, as expected, but is the most
seed-stable since $\alpha_i$ is shaped directly by the localisation loss. The
top-$k$ score~\eqref{eq:abs_topk} matches~\eqref{eq:abs_strength} to within seed
noise ($<\!0.003$ AUROC), confirming that absence is signalled by a distributed
lack of evidence rather than a single patch. We therefore adopt the
simplest, \eqref{eq:abs_strength}.

\begin{table}[t]
\centering
\caption{Absence score comparison on gRefCOCO. PROVE results are
mean$\pm$std over five seeds; baselines are single runs. The threshold
$\tau$ is selected on validation with $T\text{-acc}\!\geq\!0.90$.
Best result per split and metric is shown in \textbf{bold}.}
\label{tab:supp_absence}
\setlength{\tabcolsep}{3.5pt}
\renewcommand{\arraystretch}{1.15}
\footnotesize
\begin{tabular}{@{}clcccc@{}}
\toprule
\textbf{Split} & \textbf{Absence score} & \textbf{AUROC}
& \textbf{$N$-acc} & \textbf{$T$-acc} & \textbf{Bal.} \\
\midrule
\multirow{5}{*}{\shortstack{testA\\(people)}}
& Grad-ECLIP $-\max_i \tilde{s}_i$
& $0.682$ & $0.570$ & $0.711$ & $0.640$ \\
& Det.\ control $-\max_i \mu_i^{\mathrm{det}}$
& $0.823$ & $0.643$ & $0.864$ & $0.754$ \\
& PROVE $-\max_i \alpha_i$
& $0.829{\scriptstyle\,\pm0.002}$
& $0.644{\scriptstyle\,\pm0.003}$
& $0.883{\scriptstyle\,\pm0.006}$
& $0.763{\scriptstyle\,\pm0.002}$ \\
& PROVE $-\mathrm{top}\text{-}k\, s_i$
& $0.847{\scriptstyle\,\pm0.010}$
& $\mathbf{0.661}{\scriptstyle\,\pm0.024}$
& $0.921{\scriptstyle\,\pm0.013}$
& $0.791{\scriptstyle\,\pm0.006}$ \\
& PROVE $-\max_i s_i$
& $\mathbf{0.849}{\scriptstyle\,\pm0.010}$
& $0.657{\scriptstyle\,\pm0.029}$
& $\mathbf{0.928}{\scriptstyle\,\pm0.017}$
& $\mathbf{0.792}{\scriptstyle\,\pm0.006}$ \\
\midrule
\multirow{5}{*}{\shortstack{testB\\(objects)}}
& Grad-ECLIP $-\max_i \tilde{s}_i$
& $0.746$ & $0.518$ & $0.853$ & $0.686$ \\
& Det.\ control $-\max_i \mu_i^{\mathrm{det}}$
& $0.789$ & $0.671$ & $0.749$ & $0.710$ \\
& PROVE $-\max_i \alpha_i$
& $0.801{\scriptstyle\,\pm0.002}$
& $\mathbf{0.674}{\scriptstyle\,\pm0.006}$
& $0.796{\scriptstyle\,\pm0.007}$
& $0.735{\scriptstyle\,\pm0.002}$ \\
& PROVE $-\mathrm{top}\text{-}k\, s_i$
& $0.811{\scriptstyle\,\pm0.006}$
& $0.672{\scriptstyle\,\pm0.021}$
& $0.861{\scriptstyle\,\pm0.015}$
& $0.767{\scriptstyle\,\pm0.004}$ \\
& PROVE $-\max_i s_i$
& $\mathbf{0.813}{\scriptstyle\,\pm0.006}$
& $0.669{\scriptstyle\,\pm0.026}$
& $\mathbf{0.869}{\scriptstyle\,\pm0.021}$
& $\mathbf{0.769}{\scriptstyle\,\pm0.004}$ \\
\bottomrule
\end{tabular}
\end{table}

\begin{table}[h]
\centering
\caption{Effect of evidence capacity $\gamma$ on gRefCOCO. Best per row in \textbf{bold}.}
\label{tab:supp_gamma}
\setlength{\tabcolsep}{4pt}
\renewcommand{\arraystretch}{1.15}
\scriptsize
\begin{tabular}{@{}clccccc@{}}
\toprule
\textbf{Split} & \textbf{Metric}
& $\gamma=5$
& $\gamma=10$
& $\bm{\gamma=20}$
& $\gamma=40$
& $\gamma=100$ \\
\midrule
\multirow{4}{*}{\shortstack{testA\\(people)}}
& PG
& $0.858$ & $\mathbf{0.859}$ & $0.857$ & $0.856$ & $0.854$ \\
& EBPG
& $0.374$ & $0.423$ & $0.463$ & $0.481$ & $\mathbf{0.504}$ \\
& IoU@$k$
& $\mathbf{0.619}$ & $0.617$ & $0.616$ & $0.615$ & $0.615$ \\
& AUROC
& $0.823$ & $\mathbf{0.849}$ & $0.836$ & $0.847$ & $0.833$ \\
\midrule
\multirow{4}{*}{\shortstack{testB\\(objects)}}
& PG
& $0.759$ & $\mathbf{0.761}$ & $0.755$ & $0.757$ & $0.757$ \\
& EBPG
& $0.340$ & $0.380$ & $0.415$ & $0.427$ & $\mathbf{0.449}$ \\
& IoU@$k$
& $\mathbf{0.540}$ & $0.537$ & $0.534$ & $0.536$ & $0.537$ \\
& AUROC
& $0.799$ & $0.810$ & $0.803$ & $\mathbf{0.813}$ & $0.800$ \\
\bottomrule
\end{tabular}
\end{table}

\section{Evidence Capacity}
\label{supp:gamma}
The capacity $\gamma$ bounds the per-patch evidence through
$e_i^{\pm}=\gamma\tanh(\operatorname{softplus}(\hat e_i^{\pm})/\gamma)$, so that
$\alpha_i,\beta_i\in[1,1{+}\gamma)$, the concentration satisfies
$s_i\in[2,2{+}2\gamma)$, and the minimum attainable vacuity is $1/(1{+}\gamma)$. It thus sets the scale of the evidence rather than the relevance itself, and the main paper treats it as a capacity parameter to be studied empirically. We sweep
$\gamma\in\{5,10,20,40,100\}$, training one model per value with all other settings fixed. Table~\ref{tab:supp_gamma} reports localisation and absence
on gRefCOCO testA and testB. PG, IoU@$k$ and absence AUROC are all stable across
$\gamma$ on both splits. The only systematic trend is EBPG, which increases with $\gamma$ as the larger capacity lets the mean concentrate the relevance mass more tightly on the referent. PROVE is thus insensitive to $\gamma$ over more
than an order of magnitude, and the default $\gamma{=}20$ lies in this stable region.

\clearpage
\section{Additional Qualitative Results}
\label{sec:supp_qual}

\begin{figure*}[htbp]
    \centering
    \includegraphics[width=0.7\linewidth]{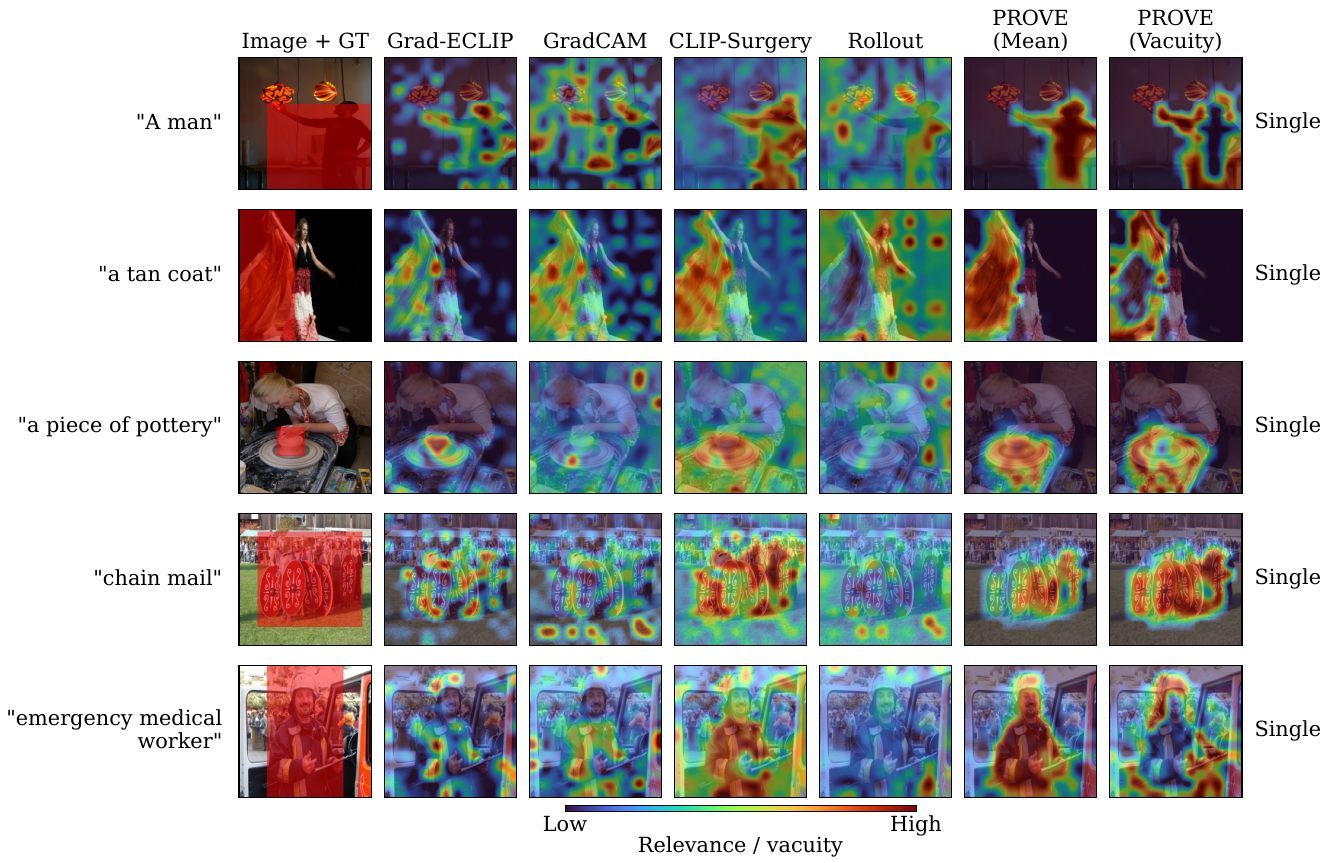}
    \caption{\textbf{Qualitative results on Flickr30k Entities.} }
    \label{fig:supp_qual}
\end{figure*}

\begin{figure*}[htbp]
    \centering
    \includegraphics[width=0.7\linewidth]{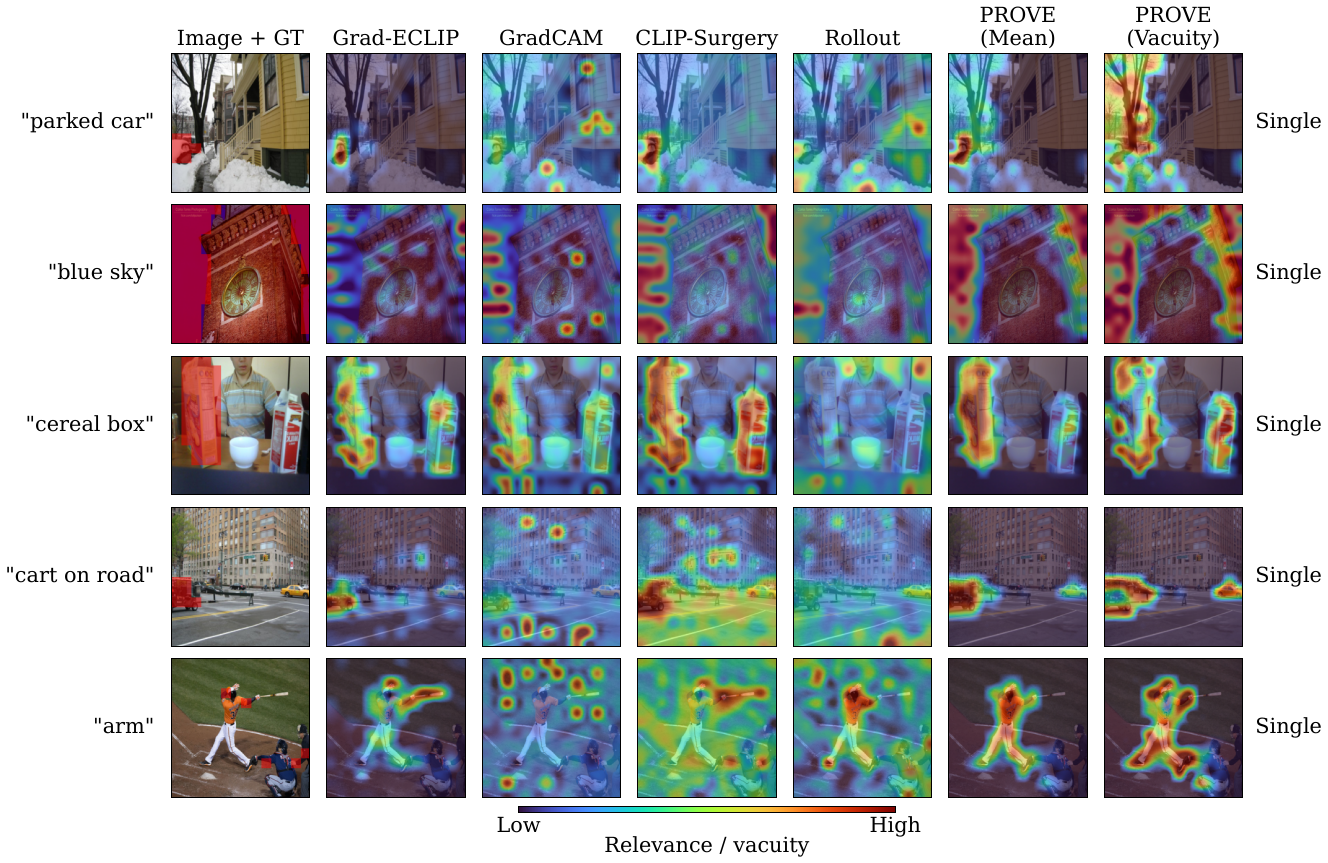}
    \caption{\textbf{Qualitative results on PhraseCut.}}
    \label{fig:supp_qual_phrasecut}
\end{figure*}

\begin{figure*}[htbp]
    \centering
    \includegraphics[width=0.8\linewidth]{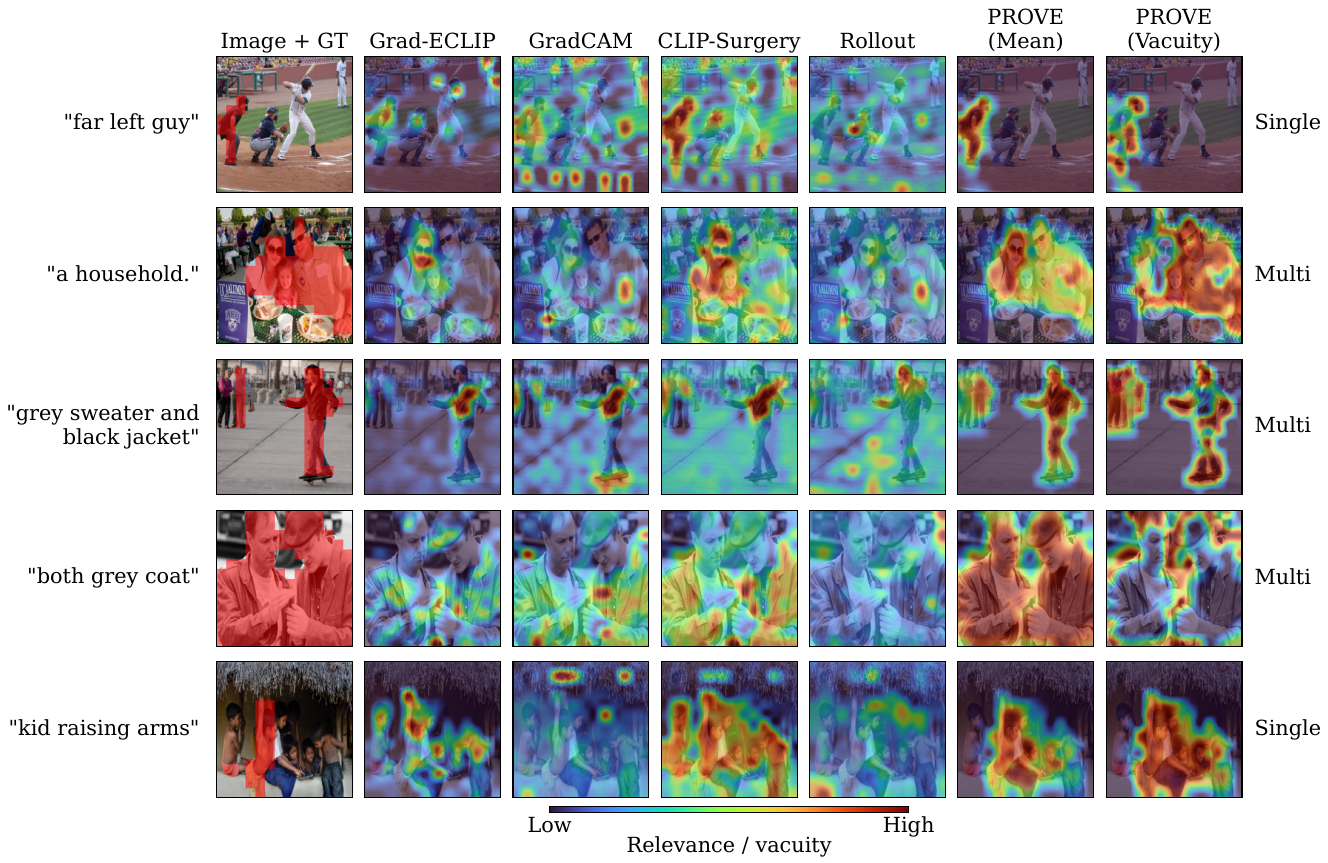}
    \caption{\textbf{Qualitative results on gRefCOCO TestA (People).}}
    \label{fig:qual_refcoco_people}
\end{figure*}

\begin{figure*}[htbp]
    \centering
    \includegraphics[width=0.8\linewidth]{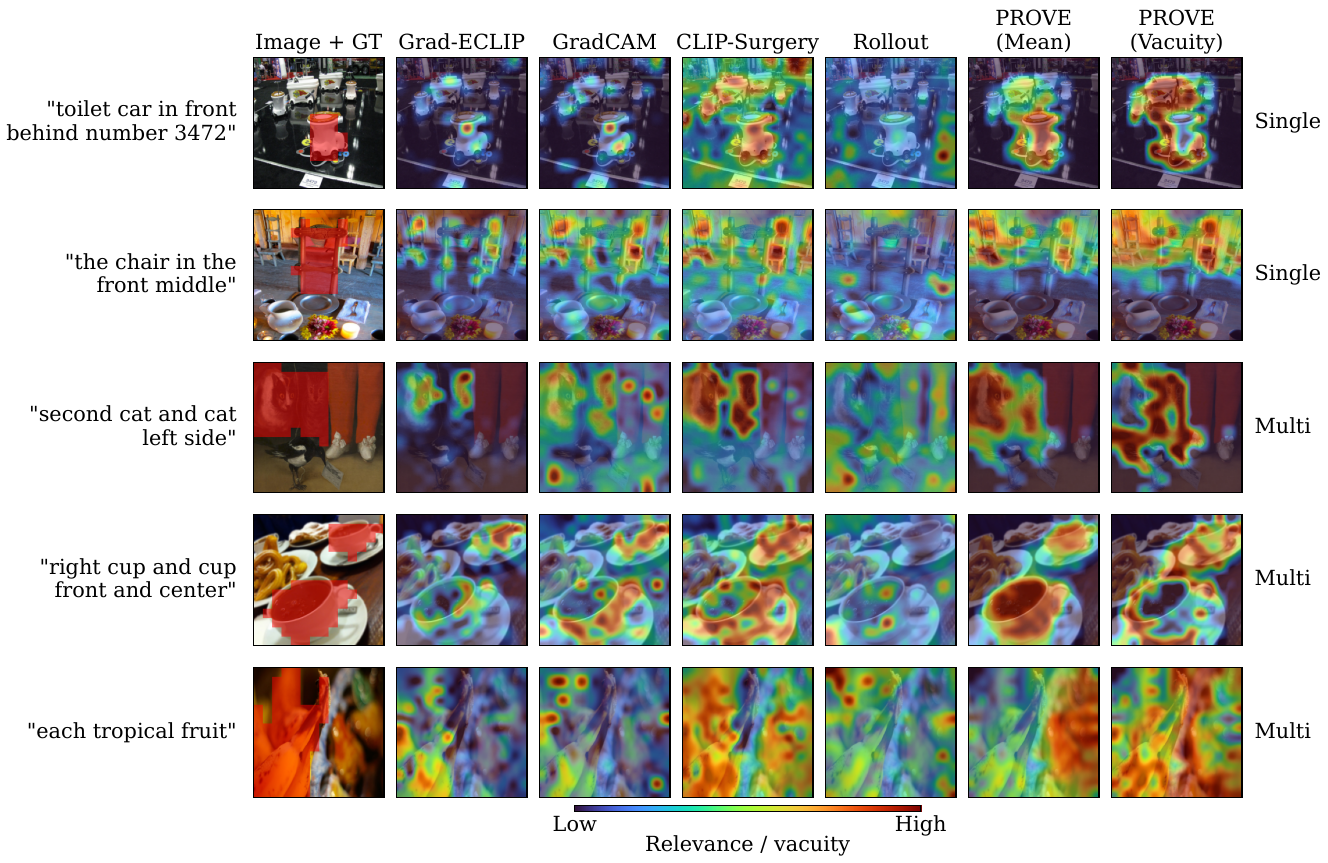}
    \caption{\textbf{Qualitative results on gRefCOCO TestB (Objects).}}
    \label{fig:qual_refcoco_objects}
\end{figure*}

Figures~\ref{fig:supp_qual}--\ref{fig:qual_refcoco_objects} show additional
qualitative examples on Flickr30k Entities, PhraseCut, and gRefCOCO testA and testB.
Each row shows the input image with the ground truth, Grad-ECLIP, GradCAM,
CLIP-Surgery, Rollout, the PROVE mean map, and the PROVE vacuity map. Vacuity is highest at object boundaries and in scenes with multiple plausible referents, where evidence is weakest, reflecting
genuine spatial ambiguity in the predicted distribution.

\end{document}